\documentclass[twoside]{article}

\usepackage[accepted]{aistats2019}

\usepackage[utf8]{inputenc} 
\usepackage[T1]{fontenc}    
\usepackage{hyperref}       
\usepackage{url}            
\usepackage{booktabs}       
\usepackage{amsfonts}       
\usepackage{nicefrac}       
\usepackage{microtype}      
\usepackage{amsmath,amsthm}
\usepackage{latexsym}
\usepackage{amssymb}
\usepackage{xspace}
\usepackage{algorithm}
\usepackage{algorithmic}
\usepackage{adjustbox}
\usepackage[capitalise]{cleveref}

\usepackage[round]{natbib}

\usepackage{xcolor}

\newtheorem{Def}{Definition}
\newtheorem{Thm}{Theorem}
\newtheorem{Lem}{Lemma}

\newtheorem*{Lem*}{Lemma}
\newcommand{\E}{\ensuremath{\mathbb{E}}}

\newcommand{\Xc}{\mathcal{X}}
\newcommand{\Yc}{\mathcal{Y}}
\newcommand{\topk}{\textstyle \text{Top}_{k}}

\newcommand\mH{\mathcal{H}}

\newcommand\R{\mathbb{R}}
\newcommand\ind{\mathbb{I}}

\newcommand\Rad{\mathfrak{R}}
\newcommand\Gauss{\mathfrak{G}}
\newcommand\ellsnm{\ell_{\text{snm}}}

\newcommand{\lshtc}{{\sc WikiLSHTC}}
\newcommand{\amsmall}{{\sc Amazon670K}}
\newcommand{\amlarge}{{\sc Amazon3M}}
\newcommand{\amcat}{{\sc AmazonCat}}

\begin{document}

%

%
\runningauthor{Reddi, Kale, Yu, Holtmann-Rice, Chen, Kumar}

\twocolumn[

\aistatstitle{Stochastic Negative Mining for Learning with Large Output Spaces}

\aistatsauthor{Sashank J. Reddi$^\dagger$, Satyen Kale$^\dagger$, Felix Yu$^\dagger$, Dan Holtmann-Rice$^\dagger$}
\aistatsauthor{Jiecao Chen*, Sanjiv Kumar$^\dagger$}
\aistatsaddress{$^\dagger$Google Research, NY \And *Indiana University, IN} ]

\begin{abstract}
We consider the problem of retrieving the most relevant labels for a given input when the size of the output space is very large. Retrieval methods are modeled as set-valued classifiers which output a small set of classes for each input, and a mistake is made if the label is not in the output set. Despite its practical importance, a statistically principled, yet practical solution to this problem is largely missing. To this end, we first define a family of surrogate losses and show that they are {\em calibrated} and {\em convex} under certain conditions on the loss parameters and data distribution, thereby establishing a statistical and analytical basis for using these losses. Furthermore, we identify a particularly intuitive class of loss functions in the aforementioned family and show that they are amenable to practical implementation in the large output space setting (i.e. computation is possible without evaluating scores of all labels) by developing a technique called {\em Stochastic Negative Mining}. We also provide generalization error bounds for the losses in the family. Finally, we conduct experiments which demonstrate that Stochastic Negative Mining yields benefits over commonly used negative sampling approaches.
\end{abstract}

\section{INTRODUCTION}

Recently, machine learning problems with extremely large output spaces have become ubiquitous: for example, extreme multiclass or multilabel classification problems with many classes, language modeling with big vocabularies, etc. Information retrieval tasks such as retrieving the most relevant documents for a given query can also be viewed as machine learning problems of this type. 

In this paper we specifically consider retrieval tasks where the objective is to output the $k$ most relevant classes for an input out of a very large number of possible classes. Training and test examples consist of pairs $(x, y)$ where $x$ represents the input and $y$ is \emph{one} class that is relevant for it. This setting is common in retrieval tasks: for example, $x$ might represent a search query, and $y$ a document that a user clicked on in response to the search query. The goal is to learn a set-valued classifier that for any input $x$ outputs a set of $k$ classes that it believes are most relevant for $x$, and the model is evaluated based on whether the class $y$ is captured in these $k$ classes.

Typically, machine learning models for such problems take the form of scoring functions that assign a real valued score to each possible output class for a given input indicating how relevant it is to the input, and outputting the classes with top $k$ scores. Such models are trained using standard loss  functions for training classifiers such as softmax cross entropy loss, max margin loss, etc. The key challenge here is that evaluating such losses requires computing the scores of all possible classes.  When the number of classes becomes very large (say, in the order of hundreds of thousands or more), this makes training as well as inference very expensive.

A variety of techniques have been developed to deal with this problem such as sampled softmax \citep{sampled-softmax}, negative sampling \citep{negative-sampling}, tree based approaches \citep{tree-based}, etc. Many of these approaches are designed for the softmax cross entropy loss, and while these methods exhibit good practical performance in few cases, many of them are biased and may not converge to the optimal solution in the limit (some notable exceptions are \citep{raman,iyengar}). For the retrieval problem that we consider in this paper, to the best of our knowledge, there is no analysis of such methods for their statistical validity for the problem. Motivated by class ambiguity in image classification tasks, \citet{hein} considered the same retrieval problem as in this paper, and designed the top-$k$ multiclass SVM algorithm for it by defining the top-$k$ hinge loss function, which unfortunately does not scale to extremely large output spaces since computing it requires computing scores of all labels.

All the aforementioned methods suffer from either statistical or scalability issues. Most practical works on large-output often resort to some variant of {\em negative sampling} due to its simplicity. Negative sampling approaches randomly sample a few classes other than the given positive class and treat them as negative classes, possibly with some additional correction, while computing the loss \citep{sampled-softmax,negative-sampling}. Statistical performance issues with negative sampling approach have recently motivated the use of some heuristics based on mining negatives with large scores, broadly referred to semi-hard negative mining \citep{facenet, tapas}. While such approaches improve the performance in some cases, they are not statistically grounded. In this paper, we develop a statistically sound and scalable approach based on the principle of mining scores from few randomly sampled negatives, and provide theoretical and empirical basis for using it over standard negative sampling approaches.

To this end, we begin with designing loss functions for the retrieval problem that have desirable statistical properties. In particular, we define a family of loss functions called \emph{Ordered Weighted Losses} (OWLs). We provide a statistical analysis of these loss functions and show that they satisfy several desirable properties under mild conditions. Furthermore, we provide a negative mining approach, called \emph{Stochastic Negative Mining}, to efficiently optimize an instance of OWLs. More specifically, our contributions are the following:
\begin{enumerate}
	\item {\bf Calibration.} We show OWLs are calibrated and therefore training models by minimizing OWLs leads to the Bayes optimal predictor in the limit as the model capacity increases and the number of samples grows.
	
	\item {\bf Convexity.} We show that OWLs are convex in the score vector. Thus for linear and kernel models for computing scores, minimizing the training loss is a convex problem in the model parameters and thus standard convex optimization techniques can be used.
	
	\item {\bf Surrogate loss.} We show that OWLs are valid surrogate losses for the $0/1$ retrieval loss of interest.
	
	\item {\bf Practical implementability.} We provide an instance of OWLs that can be efficiently optimized through Stochastic Negative Mining. This technique samples a set of classes and treats the highest scoring ones in the set as ``negative'' classes for the training example; thereby, avoiding evaluating scores of all labels. This technique has intuitive appeal and has been empirically observed to yield good performance since it avoids computing scores of all the possible output classes.
	
	\item {\bf Generalization error bounds.} We provide generalization error bounds for OWLs that provide guidance on how to choose OWL parameters. We also provide margin bounds for the retrieval loss for arbitrary hypothesis classes in terms of their Gaussian and Rademacher complexities.
	
	\item {\bf Experimental validation.} We provide experimental evidence that Stochastic Negative Mining does indeed help improve performance when learning with large output spaces compared to simpler sampling based strategies.
\end{enumerate}

\subsection{Related work}
There is extensive literature on the problem of learning set-valued classifiers: see, for example, \cite{grycko,delcoz,vgs05,wlw04,lrw13}, particularly in the context of {\em binary classification with a reject option} \citep{chow70,hw06,bw08,yw10}. \cite{delcoz} specifically considered using such classifiers in the information retrieval context. \cite{dh17} considered a similar setting to the one in this paper and provided a procedure with bounded expected size of the output set and establish statistical optimality of the procedure. \cite{slw18} provide characterizations of optimal set-valued classifiers with user-defined levels of coverage or confidence and estimators with good asymptotic
and finite sample properties. 

Another related setting is the \emph{learning to rank} problem (see e.g. \cite{joachims,agarwal,boyd,kar} and the references therein). The objective there is to rank a given set of items so that relevant items are ranked as highly as possible. Training data consist of items along with a binary label indicating whether the item is relevant or not. Various performance metrics are considered such as Precision@$k$ (the fraction of the top $k$ ranked items that are relevant), the normalized discounted cumulative gain (NDCG) and other variants of DCG, or the mean reciprocal rank (MRR), etc. 

The learning to rank setting is different than ours, however. One can view the setting in this paper as a more general \emph{contextual} version of the learning to rank problem, with the added difficulty that we do not get to observe \emph{irrelevant} classes for our inputs. Our goal here is to study retrieval methods that rank classes for each input context so as to maximize the fraction of relevant classes that land in the top-$k$ ranked classes. The performance metric we consider is closely related to the Recall@$k$ in the learning to rank setting: specifically, if the distribution on relevant classes conditioned on any given input is uniform, then the metric is exactly the expected Recall@$k$ over randomly chosen inputs. However we emphasize that we do not restrict the class conditional distribution to be uniform.

A closely related work to ours in the learning to rank literature is that of \citet{kar} who studied the Precision@$k$ metric and provided surrogate loss functions for it and showed calibration under various conditions, gave methods for efficiently optimizing them, and provided generalization bounds. Another closely related work is that of \cite{ubg09} who also developed loss functions that are essentially the same as the Pairwise Ordered Weighted Losses in this paper. Turning to the optimization, the seminal work of \citet{joachims} gave an SVM method to optimize ranking metrics including Recall@$k$. However this method does not scale to large datasets since the loss function used is not decomposable. \citet{global-objectives} gave convex relaxations for information retrieval metrics with decomposable objectives, leading to training that scales to large datasets, although the relaxations do not come with theoretical guarantees other than being valid surrogates for the ranking metric in question. 

None of the above works specifically tackle the problem of learning with very large output spaces that we consider in this paper. To the best of our knowledge, there has been no prior work on calibrated surrogate losses for the 0/1 retrieval loss we consider in this paper.

\section{PRELIMINARIES AND NOTATION}

The input space for examples is denoted $\Xc$ and the output space $\Yc$. Define $K := |\Yc|$, and we identify the output classes in $\Yc$ with the integers $1, 2, \ldots, K$. The standing assumption in this paper is that $K$ is ``large'': in the tens of thousands or larger. Unless specified otherwise, all vectors live in $\R^K$, and $\|v\|_p$ denotes the $p$-norm of $v$. For any vector $v \in \mathbb{R}^K$, we use $v_{[i]}$ to denote the $i^{th}$ largest element of $v$ and $\topk(v)$ to denote indices of the largest $k$ coordinates in the vector $v$ (breaking ties arbitrarily). For any $y \in \Yc$, we use $v^{-y}$ to denote the vector in $\R^{K-1}$ obtained by dropping coordinate $v_y$ from $v$. We use $\mathbb{I}$ to denote the indicator function: $\ind(x) = 1$ if $x$ is true, and $0$ otherwise. For any $m \in \mathbb{N}$, $[m] := \{1, 2, \ldots, m\}$.

We consider the following retrieval problem. There is an unknown distribution $D$ over $\Xc \times \Yc$. A sample $(x, y)$ drawn from this distribution indicates that $y$ is a relevant class for $x$. For a parameter $k \in \mathbb{N}$ with $k \ll K$, the goal is to design a method that, given an input $x \in \Xc$, outputs the $k$ ``most relevant'' classes $y \in \Yc$. 
Formally, a retrieval method, also called a predictor, is a function that for any input outputs a set of labels of size $k$, i.e. a function $f: \Xc \rightarrow \Yc_k$ where $\Yc_k = \{\mathcal{U} \subseteq \Yc \mid |\mathcal{U}| = k\}$. 
Let $L: \Yc_k \times \Yc \rightarrow \R$ be a loss function that measures quality of the predictor's output: $L(S, y)$ should be large if $y \not\in S$.  
In the retrieval setting of our interest, a commonly used loss function is the following \emph{$0/1$ retrieval loss}:
\begin{equation}
	L(S, y) = \mathbb{I}(y \notin S).
	\label{eq:true-loss}
\end{equation}
Our aim is to find a predictor $f$ with low expected loss (also referred to as {\em risk}):
\[R(f) := \mathbb{E}_{(X,Y) \sim D}[L(f(X), Y)].\] 
Note that for $k=1$, this effectively reduces to the classical multiclass classification problem. 

One simple observation is the following characterization of the Bayes optimal predictor:
\begin{Lem}
	The predictor $b(x) := \topk [D(Y=1|X=x), \cdots, D(Y=K|X=x))]$ has minimal risk over all predictors.
\end{Lem}
\begin{proof}
	For any predictor $f: \Xc \rightarrow \Yc_k$, we have $R(f) = \mathbb{E}_X \left[ \sum_{y \in \Yc \backslash f(X)} D(Y = y \mid X) \right]$. This is minimized by $f = b$.
\end{proof}

Without loss of generality we assume that $D_{[k]}(\cdot\mid X = x) > D_{[k+1]}(\cdot\mid X = x)$ for all $x \in \Xc$, so that $b$ is uniquely defined.\footnote{The results in this paper can be easily applied to the case when $D_{[k]}(\cdot\mid X = x) = D_{[k+1]}(\cdot\mid X = x)$.}

Inspired by the above characterization of the Bayes optimal predictor, as well as standard practice, we aim to learn a predictor by finding a scoring function $h : \Xc \rightarrow \mathbb{R}^K$ and predict the set of labels of $x$ as $f(x) = \topk(h(x))$.

\begin{Def}
	We say a score vector $v \in \R^K$ is Bayes compatible for $x \in \Xc$ if $\topk(v) = b(x)$.
\end{Def}
We denote the set of all Bayes compatible vectors
for any $x \in \Xc$ by $\mathcal{B}_x$. Ideally, we would like to learn a hypothesis which outputs a Bayes compatible vector for each $x \in \Xc$. Since minimizing the $0/1$ loss is typically intractable, in practice we instead use a more tractable {\em surrogate loss} $\ell: \R^K \times \Yc \rightarrow \R$. Define $\ell$-risk as $R_\ell(h) := \mathbb{E}_{(X, Y) \sim D}[\ell(h(X), Y)]$. Given a training set of $n$ samples $\{(x_i, y_i)\}_{i=1}^n$ drawn i.i.d. from $D$, the empirical $\ell$-risk is $\widehat{R}_\ell(h) := \frac{1}{n} \sum_{i=1}^n \ell(h(x_i), y_i)$. A predictor is then computed by minimizing the empirical $\ell$-risk over an appropriate hypothesis class $\mH$.


\newtheorem{Assump}{Assumption}
\newcommand\sumtop{\mathrm{sumtop}}


A basic and desirable property of the loss function is \emph{calibration}, which is defined below.
\begin{Def}
We say loss function $\ell$ is calibrated with respect to the retrieval loss \eqref{eq:true-loss} if the following condition holds for all $x \in \mathcal{X}$:
\begin{align}
\inf_v \mathbb{E}_{Y}[\ell(v, Y) \mid X=x]  < \inf_{v \notin \mathcal{B}_x} \mathbb{E}_{Y}[\ell(v, Y) \mid X=x] .
\end{align}
\label{def:calibration}
\end{Def}
The above definition is a natural generalization of the one in \cite{Zhang04}. This definition essentially states that the loss function is calibrated if, given any particular input $x$, score vectors that minimize the loss function are Bayes compatible for $x$. The following result (proof in Appendix~\ref{sec:asymptotic-convergence-proof}) shows that minimizing a calibrated surrogate loss leads to the Bayes optimal predictor:
\begin{Thm}
 \label{thm:calibration}
  Let $\mH$ be the class of all measurable functions $h: \Xc \rightarrow \R^K$. If $\ell$ is calibrated, then for all $\epsilon > 0$, there exists $\delta > 0$ such that for any distribution $D$ and any $h \in \mH$, if $R_\ell(h) \leq \inf_{h' \in \mathcal{H}} R_\ell(h') + \delta$, then 
  $R(\topk(h)) \leq \inf_{h' \in \mathcal{H}}R(\topk(h')) + \epsilon$.
\end{Thm}

\section{ORDERED WEIGHTED LOSSES AND STOCHASTIC  NEGATIVE MINING}
We now present some loss functions specifically geared towards scenarios where $K$ is large. We start by defining a general class of loss functions (hereafter referred to as ordered weighted loss (OWL)). OWLs are parameterized by
\begin{enumerate}
	\item a non-increasing function $\phi: \R \rightarrow \R$ that is a surrogate loss for the $0/1$ loss, i.e. $\phi(x) \geq \ind(x \leq 0)$ for all $x \in \R$, and
	\item a non-negative weight vector $\theta \in \R^{K-1}$.
\end{enumerate}
Some examples of valid $\phi$ functions are the hinge loss $\phi_{\text{hinge}}(u) = \max\{1 - u, 0\}$, logistic loss $\phi_{\text{logistic}}(u) = \log_2(1 + \exp(-u))$, squared-hinge loss $\phi_{\text{sq-hinge}}(u) = \max\{1 - u, 0\}^2$, 
exponential loss $\phi_{\text{exp}}(u) = \exp(-u)$, and ramp loss, parameterized by a margin $\rho > 0$: $\phi_{\text{ramp},\rho}(u) = \ind(u \leq 0) + (1 - u/\rho)\ind(0 < u \leq \rho)$.

We define two types of OWLs:
\begin{Def} \label{def:snm-pairwise-loss-alt}
The $(\phi, \theta)$-{\bf Pairwise} Ordered Weighted Loss (POWL) $\ell$ is defined as 
$\ell(v, y) := \sum_{j=1}^{K-1} \theta_j \phi(v_y~-~v^{-y}_{[j]})$.
\end{Def}
\begin{Def} \label{def:snm-binary-loss-alt}
The $(\phi, \theta)$-{\bf Binary} Ordered Weighted Loss (BOWL) $\ell$ is defined as
$\ell(v, y) := \phi(v_y) + \sum_{j=1}^{K-1} \theta_j \phi(-v^{-y}_{[j]})$.
\end{Def}

POWLs have also been studied by \cite{ubg09}. Several commonly used loss functions for multiclass classification are OWLs: for example, the multiclass SVM loss of \cite{crammer-singer} is a POWL with $\theta_1 = 1$ and $\theta_j = 0$ for all $j > 1$, the loss function of \cite{weston-watkins} is a POWL with all $\theta_j = 1$, and the loss function of \cite{lee-lin-wahba} is a BOWL with all $\theta_j = 1$.

With the notable exception of the ramp loss $\phi_{\text{ramp},\rho}$, all other examples for $\phi$ are convex. For convex $\phi$, under some mild additional conditions on $\theta$, we can show that the two types of OWLs defined above are convex in the score vector $v$. If $v$ is generated via a linear or kernel model, the loss becomes convex in the model paramters and can be optimized using convex optimization techniques.
\begin{Thm}[Convexity] \label{thm:convexity}
Suppose $\phi$ is convex. Furthermore, suppose that $\theta$ has non-increasing coordinates, i.e. $\theta_j \geq \theta_{j'}$ for $j < j'$. 
Then both the $(\phi, \theta)$-POWL and the $(\phi, \theta)$-BOWL are convex in the score vector $v$ for any fixed $y \in \Yc$.
\end{Thm}
The proof appears in Appendix~\ref{sec:convexity-surrogate-proofs}. Furthermore, under a different condition on $\theta$, the OWLs are valid surrogate losses for the retrieval loss~\eqref{eq:true-loss} (proof in Appendix~\ref{sec:convexity-surrogate-proofs}):
\begin{Thm}[Surrogate loss]
Suppose $\theta_j = \nicefrac{1}{k}$ for all $j \in [k]$. Then the $(\phi, \theta)$-POWL $\ell$ is a surrogate loss for the retrieval loss~\eqref{eq:true-loss}: $\ell(v, y) \geq \ind(y \not\in \topk(v))$. Similarly, the $(\phi, \theta)$-BOWL $\ell$ is a surrogate loss for the retrieval loss scaled by $2$: $\ell(v, y) \geq 2\ind(y \not\in \topk(v))$.
\label{thm:surrogate}
\end{Thm}

Finally, under certain conditions on $\phi$, $\theta$, and the data distribution, OWLs are calibrated:
\begin{Thm}[Calibration]
Suppose $\phi$ is a differentiable function with $\phi'(\epsilon) < 0$ for $\epsilon \leq 0$, and $\theta$ is such that $\theta_i = \tfrac{1}{k}$ for $j \in [k]$ and $\theta_j \leq \tfrac{1}{k}$ for $j > k$. Also, suppose the following condition holds for all $x \in \Xc$:
\begin{equation} \label{eq:calibration-condition}
D_{[k]}(\cdot \mid X=x) > \frac{\sum_{l=k+1}^{m} D_{[l]}(\cdot \mid X=x))}{k \sum_{j=k+1}^{m} \theta_{j}},	
\end{equation}
for all $m > k$. Then we have the following:
\begin{enumerate} 
\item The $(\phi, \theta)$-POWL is calibrated.
\item Suppose additionally that $\theta_j > 0$ for all $j$. Then the $(\phi, \theta)$-BOWL is calibrated.
\end{enumerate}
\label{thm:snm-calibration}
\end{Thm}
The proofs of these results appear in Appendix~\ref{sec:calibration-proof}.

The condition~\eqref{eq:calibration-condition} required to show calibration is essentially an assumption on the tail of the class conditional distribution for $x$. While it is not possible to verify the condition since we don't have access to the data distribution, it is still possible in practical applications to choose $\theta$ such that this condition holds. For example, suppose from domain knowledge we know that for any $x$, there can be at most $M \ll K$ relevant labels (i.e. $D_{[M+1]}(\cdot | X = x) = 0$). 
In image labeling tasks, for example, we may have reason to expect that any image can have no more than 10 different labels. 
In that case, we can set $\theta_j = \nicefrac{1}{k}$ for $j \in [M]$ and $\theta_j = 0$ for $j > M$ suffices to satisfy this condition. One can make a similar prescription for $\theta$ for milder domain knowledge requirements; for example in cases where we know that for any $x$, $\sum_{j=M+1}^{K-1} D_{[j]}(\cdot | X = x) < \epsilon$ for some small $\epsilon$ like $0.1$. Finally, the default setting of $\theta_j = \nicefrac{1}{k}$ for all $j \in [K-1]$ always satisfies \eqref{eq:calibration-condition}.

\subsection{Stochastic Negative Mining}

The above results show that the proposed family of loss functions has useful statistical properties. However the choice of $\theta$ and the computational efficiency of minimizing such a loss function have not been discussed, specifically for the large $K$ setting. This is central to the paper since, in our problem setting, we desire a loss function that can be computed (or at least, randomly estimated) without needing to compute the scores of all $K$ labels. For instance, one could simply choose $\theta_j = 0$ for all $j > k$, typically referred to as \emph{top-k} loss, but it is computationally intractable for very large $k$.

\begin{algorithm}[t]
     \caption{Stochastic Negative Mining}
 \begin{algorithmic}[1] \label{alg:snm-loss}
     \vspace{+0.1cm}
     \REQUIRE A score vector $v \in \R^K$ with random coordinate access, a class $y \in \Yc$, a parameter $B \in \mathbb{N}$ with $B \geq k$, a non-negative vector $\vartheta \in \R^B$ with non-increasing coordinates, and desired loss type (POWL or BOWL)
     \ENSURE A random variable $L(v, y) \in \R$ estimating the loss.
     \vspace{+0.1cm}
     \STATE Sample a subset of $\mathcal{B} \subseteq \Yc \setminus \{y\}$ of size $B$ uniformly at random.
     \STATE Let $v^{\mathcal{B}}_{[1]} \geq v^{\mathcal{B}}_{[2]} \cdots \geq v^{\mathcal{B}}_{[B]}$ be elements of $\{v_{y'}\ |\ y' \in \mathcal{B}\}$ in non-increasing order.
     \STATE Return 
     \[L(v, y) = \begin{cases}
     \sum_{j = 1}^{B} \vartheta_j\phi(v_y - v^{\mathcal{B}}_{[j]}) & \text{ (if POWL)} \\ 
     \phi(v_y) + \sum_{j=1}^{B} \vartheta_j\phi(-v^{\mathcal{B}}_{[j]}) & \text{ (if BOWL)}
     \end{cases}
     \]
 \end{algorithmic}
 \end{algorithm}

To this end, we now present an approach to efficiently optimize a particular instance of OWL; thereby enjoying the useful statistical properties described above in addition to being amenable to efficient computation. For the ease of exposition, instead of directly specifying the value of $\theta$, we resort to a constructive approach to describe the loss function. The construction samples a set $\mathcal{B}$ of $B$ labels in $\Yc \setminus \{y\}$ uniformly at random, then sorts the scores of the sampled labels, and computes the loss in an OWL-like manner using a weight vector $\vartheta \in \R^B$ (see Algorithm~\ref{alg:snm-loss}). The actual loss function is defined as $\ellsnm(v, y) := \E_{\mathcal{B}}[L(v, y)]$. 
We can obtain an unbiased estimate of the loss (and its gradients) without having to compute this expectation explicitly by simply computing the loss on a randomly selected subset $\mathcal{B}$ (as described in Algorithm~\ref{alg:snm-loss}); thus, allowing efficient optimization using algorithms like stochastic gradient method. We term this procedure \emph{Stochastic Negative Mining} (SNM).

It should be evident from the description of the procedure that the randomized estimator can be computed by computing the scores of only at most $B$ randomly chosen classes and the score of the class $y$ in question. Overall, including the time for sampling the classes, the procedure can be implemented in $O(B \log(K))$ time, which can be a significantly faster than computing scores of all labels if $B \ll K$. The tradeoff is increased variance in the estimator leading to worse generalization bounds (as we shall see shortly) compared to computing all scores, but in practice the benefits can significantly outweigh the costs, as our experiments demonstrate. 

To gain more intuition, observe that in the two extreme cases of $\mathcal{B} = [K] \backslash \{y\}$ (with $\vartheta_j = \tfrac{1}{k}$ for $j \leq k$ and $\vartheta_j = 0$ otherwise) and $|\mathcal{B}| = B$ (with $\vartheta_j = \tfrac{K-1}{k B}$ for $j \leq B$), the procedure amounts to optimizing \emph{top-k} loss and using negative sampling respectively. Thus, SNM seamlessly generalizes to various approaches used in the machine learning literature. The size of $\mathcal{B}$ is typically constrained by  computation and memory budgets. Using $\mathcal{B} = [K]$ is typically intractable for large $K$. When $\mathcal{B} = B$ (where $k < B \ll K$), we claim that it is beneficial to use SNM instead of the negative sampling procedure used while dealing with large-output spaces. This claim is backed through generalization bounds and empirical results (Sections~\ref{sec:generalization-bounds} \&~\ref{sec:experiments}).

We now show that $\ellsnm$ is an OWL, and thereby inherits all the useful statistical properties of OWLs. The proof is deferred to Appendix~\ref{sec:powl-bowl-proof}.
\begin{Lem}
For any $\tau > 0$, $\ellsnm$ is either a POWL or BOWL. Furthermore, the coordinates of the corresponding weight vector $\theta$ are in decreasing order and non-zero and $\ellsnm$ is convex. If $\vartheta_j = \frac{K-1}{kB}$ for $j \in [k]$, then $\theta_j = \nicefrac{1}{k}$ for all $j \in [k]$ and $\ellsnm$ is a valid surrogate loss for the retrieval loss. If $\vartheta_j > 0$ for all $j \in [B]$ and the conditions of Theorem~\ref{thm:snm-calibration} hold, then $\ellsnm$ is calibrated. Finally, $\|\theta\|_1 = \|\vartheta\|_1$, and $\|\theta\|_2 \leq \sqrt{\frac{B\vartheta_1\|\vartheta\|_1}{K-1}}$.
\label{lem:prob-dominance}
\end{Lem}

The bounds on the norms of $\theta$ mentioned above are important for the generalization bounds given in the next section. Smaller norms have smaller generalization error, and thus a good choice of $\vartheta$ is $\vartheta_j = \frac{K-1}{kB}$ for $j \in [k]$ and $\vartheta_j = 0$ for $j > k$\footnote{Setting $\vartheta_j = 0$ may come at the price of calibration, but we can rectify that by setting $\vartheta_j = \epsilon$ for some small $\epsilon$ for all $j > k$.}. For this setting of $\vartheta$ SNM reduces to the following intuitively appealing algorithm: sample a batch of $B$ classes, and choose the top-$k$ scoring classes as ``negatives'' and set the loss to be their average loss. For this reason, we call this \emph{Top-$k$ SNM}. Empirically this technique works quite well, as can be seen from our experiments.

The $\vartheta$ parameter allows for considerable flexibility in designing Stochastic Negative Mining. Other settings of $\vartheta$, than the one mentioned above, can be used based on the application. In our experiments, for example, we found that Top-$k'$ SNM for $k' < k$ works even better than Top-$k$ SNM. Another example is if in an application we wish to penalize harder negative even more than in Top-$k$ SNM, then we can choose the coordinates of $\vartheta$ according to a power law distribution with some exponent $\alpha$. Another idea is to treat $\vartheta$ as scaled sampling probabilities, sub-sample negatives within $\mathcal{B}$ according to these probabilites, and add up (appropriately scaled) losses for the sub-sampled negatives. This can lead to further computational gains since losses need to be evaluated for even fewer classes.

\section{GENERALIZATION ERROR BOUNDS}
\label{sec:generalization-bounds}
We now turn to generalization bounds for OWLs. To describe the bounds, we need to define some notation first. Let $\mH$ be a hypothesis class of functions $h: \Xc \rightarrow \R^K$. For a set $S$ of examples $(x, y) \in \Xc \times \Yc$, let $g = \{g_{(x, y)}\ |\ (x, y) \in S\}$ be a set of i.i.d. Gaussian random variables indexed by examples in $S$, and let $\E_g[\cdot]$ denote expectation over these random variables. Then the empirical Gaussian complexity w.r.t. $S$ is defined to be $\Gauss_S(\mH) = \frac{1}{|S|}\E_g[ \sup_{h \in \mH} \sum_{(x, y) \in S} g_{(x,y)}h_y(x)]$. Empirical Rademacher complexity $\Rad_S(\mH)$ is defined similarly with the Gaussian random variables replaced by Rademacher ones. We also define the {\em label completion} of $S$, denoted $\tilde{S} = \{(x, y)\ |\ \exists y' \in \Yc \text{ s.t. } (x, y') \in S,\ y \in \Yc\}$. The {\em worst-case} empirical Rademacher complexity over $\bar{S}$ is defined as $\overline{\Rad}_{\bar{S}}(\mH) := \sup\{ \Rad_{T}(\mH)\ |\ T \text{ is multi-subset of } \bar{S}, |T| = |\bar{S}|\}$. Finally, in this section we use the $\tilde{O}(\cdot)$ notation to suppress polylogarithmic factors in the problem parameters. The main generalization bound is the following, proved in Appendix~\ref{sec:gen-proofs}:
\begin{Thm} \label{thm:owl-generalization-bound}
Let $\phi(\cdot)$ be $L$-Lipschitz. Assume that for some $\Phi > 0$, $|\phi(h_y(x))|, |\phi(h_y(x) - h_{y'}(x))| \leq \Phi$ for any $y, y' \in \Yc$. Let $S$ be a sample set of $n$ i.i.d. examples drawn from the input distribution. Suppose $\ell$ is the $(\phi, \theta)$-POWL. Then with probability at least $1-\delta$ over the choice of $S$, for any $h \in \mH$, the generalization error $\E_{(x, y)}[\ell(h(x), y)] - \E_{(x, y) \sim S}[\ell(h(x), y)]$ is bounded by
\small
\begin{align*}
& \tilde{O}\left(L \min\left\{\|\theta\|_2K\Gauss_{\bar{S}}(\mH) + \|\theta\|_1\Gauss_S(\mH), 2\|\theta\|_1\sqrt{K}\overline{\Rad}_{\bar{S}}(\mH)\right\}\right) \\
& \qquad + 3\|\theta\|_1\Phi \sqrt{\frac{\log(2/\delta)}{2n}}.
\end{align*}
\normalsize
If $\ell$ is the $(\phi, \theta)$-BOWL, then the generalization error is bounded by
\small
\begin{align*}
&\tilde{O}\left(L \min\left\{\|\theta\|_2K\Gauss_{\bar{S}}(\mH) + \Gauss_S(\mH), (\|\theta\|_1+1)\sqrt{K}\overline{\Rad}_{\bar{S}}(\mH)\right\}\right) \\
& \qquad + 3\|\theta\|_1\Phi \sqrt{\frac{\log(2/\delta)}{2n}}.
\end{align*}
\normalsize
\end{Thm}

We can now analyze the effect of the parameter $B$ in Stochastic Negative Mining for the particular choice of $\vartheta$ given after Lemma~\ref{lem:prob-dominance}, i.e. $\vartheta_j = \frac{K-1}{kB}$ for $j \in [k]$ and $\vartheta_j = 0$ for $j > B$. The corresponding $\ellsnm$ has $\|\theta\|_1 = \nicefrac{K-1}{B}$ and $\|\theta\|_2 \leq \sqrt{\nicefrac{K-1}{kB}}$. The generalization error therefore decreases with $B$, as expected; albeit, at the cost of additional computation.

It is easy to check that the corresponding values of $\|\theta\|_1$ for SNM and negative sampling are $\tfrac{K-1}{B}$ and $\tfrac{K-1}{k}$ respectively. Our generalization bounds indicate that for $|\mathcal{B}|=B > k$, one can obtain better generalization through SNM in comparison to negative sampling. This is due to the fact that the generalization bounds depend on $\|\theta\|_1$, deteriorating as $\|\theta\|_1$ increases. Before ending our discussion, we need to make it explicit that our analysis only compares the upper bounds and hence, needs to be interpreted with caution. Nonetheless, our empirical evaluation, in the next section, supports our theoretical analysis and provides compelling case to use SNM approach in practice.

\begin{table*}[ht]
	\centering
        \small
	\begin{tabular}{ l | l | l | l | l | l | l  }
		\hline
		Dateset & \#Features & \#Labels & \#TrainPoints &\#TestPoints & Avg. \#P/L & Avg. \#L/P  \\
		\hline
		\amcat  & 203,882     & 13,330    & 1,186,239 & 306,782 & 448.57 & 5.04\\
		\hline    
		\lshtc  & 1,617,899    & 325,056   & 1,778,351 & 587,084 & 17.46 & 3.19\\
		\hline
		\amsmall& 135,909     & 670,091   & 490,449  & 153,025 & 3.99  & 5.45\\
		\hline
		\amlarge& 337,067     & 2,812,281  & 1,717,899 & 742,507 & 31.64 & 36.17\\
		\hline
	\end{tabular}
	\caption{Summary of the datasets used in the paper. \#P/L is the number of points per label, and \#L/P is the number of labels per point.}\label{tb:datasets}
\end{table*}

\begin{table*}[ht]
	\centering
        \small
	\begin{tabular}{ccccccc}
		& & Top 1 & Top 16 & Top 64 & Top 256 \\
		\hline
		& R@1 & 2.59 & 2.02 & 1.65 & 1.32 \\
		\amcat & R@3 & 1.98 & 1.97 & 1.63 & 1.32 \\
		& R@5 & 2.58 & 1.96 & 1.60 & 1.29 \\
		\hline
		& R@1 & 2.53 & 2.35 & 2.13 & 1.87 \\
		\lshtc & R@3 & 2.71 & 2.46 & 2.18 & 1.86 \\
		& R@5 & 2.64 & 2.37 & 2.14 & 1.83\\
		\hline
		& R@1 & 1.23 & 1.17 & 1.13 & 1.11 \\
		\amsmall & R@3 & 1.28 & 1.23 & 1.17 & 1.15\\
		& R@5 & 1.32 & 1.24 & 1.18 & 1.15 \\
		\hline
		& R@1 & 2.60 & 2.56 & 2.30 & 1.93\\
		\amlarge & R@3 & 2.92 & 2.72 & 2.42 & 2.05\\
		& R@5 & 3.01 & 2.80 & 2.51 & 2.13 \\ \hline
	\end{tabular}
        \begin{tabular}{ccccccc}
		& & Top 1 & Top 16 & Top 64 & Top 256 \\
		\hline
		& P@1 & 2.33 & 1.99 & 1.66 & 1.30 \\
		& P@3 & 2.40 & 1.97 & 1.65 & 1.30 \\
		& P@5 & 2.39 & 1.92 & 1.61 & 1.30 \\
		\hline
		& P@1 & 2.56 & 2.36 & 2.17 & 1.89 \\
		& P@3 & 2.70 & 2.46 & 2.18 & 1.90\\
		& P@5 & 2.59 & 2.37 & 2.17 & 1.87 \\
		\hline
		& P@1 & 1.25 & 1.21 & 1.16 & 1.13 \\
		& P@3 & 1.27 & 1.22 & 1.18 & 1.14 \\
		& P@5 & 1.33 & 1.34 & 1.23 & 1.18\\
		\hline
		& P@1 & 2.73 & 2.57 & 2.34 & 1.96\\
		& P@3 & 2.95 & 2.80 & 2.47 & 2.13\\
		& P@5 & 3.07 & 2.89 & 2.57 & 2.11 \\ \hline
	\end{tabular}
	\caption{Comparison of stochastic negative mining using top-$k$ SNM for various values of $k$ with negative sampling. For the ease of comparison, each entry in the table represents the Recall@$k$ (resp. Precision@$k$) value of the method normalized with the Recall@$k$ (resp. Precision@$k$) obtained for negative sampling. Note that values larger than 1 indicate better performance in comparison to negative sampling.}
	\label{tb:topk}
\end{table*}

\subsection{Margin bounds for retrieval loss}
We now provide margin based generalization error bounds for predicting labels by taking $\topk(h(x))$ for $h \in \mH$. For a hypothesis $h \in \mH$ and an example $(x, y)$, we define a notion of margin as $\rho_h(x, y) = h_y(x) - h_{[k]}^{-y}(x)$. In the multiclass setting, i.e. $k = 1$, this reduces to the standard definition of margin \citep{koltchinskii-panchenko}. Note that for any example $(x, y)$, $y \not\in \topk(h(x)) \Leftrightarrow \rho_h(x, y) \leq 0$ \footnote{There's a subtlety here in the handling of ties at the $k$-th largest score. If there's a tie, then none of the tied classes are considered valid. This is consistent with previous definitions of the margin.}. Let $S$ be a set of $n$ labeled examples drawn i.i.d. from the input distribution. We define the margin $\rho$ empirical risk of a hypothesis $h$ as $\hat{R}_{S, \rho}(h) := \E_{(x, y) \sim S}\left[ \ind(\rho_h(x, y) \leq \rho)\right]$.
With these definitions the following margin bound (proved in Appendix~\ref{sec:gen-proofs}):
\begin{Thm} \label{thm:margin-bound} Fix any $\rho > 0$.  Then with probability at least $1-\delta$, for any $h \in \mH$, we have
\begin{align}
R(h) & \leq \hat{R}_{S, \rho}(h) + 3\sqrt{\frac{\log(2/\delta)}{2n}} \nonumber \\
& \quad + \tilde{O}\left(\tfrac{1}{\rho}\min\left\{K\Gauss_{\tilde{S}}(\mH) + \Gauss_S(\mH), \sqrt{K}\bar{\Rad}_{\tilde{S}}(\mH)\right\}\right) \nonumber.
\end{align}
\end{Thm}

\section{EXPERIMENTS}
\label{sec:experiments}

We now present empirical results for the SNM approach. We use publicly available ``extreme multilabel classification'' datasets for all our experiments \footnote{The datasets are available at \url{http://manikvarma.org/downloads/XC/XMLRepository.html}} (see Table \ref{tb:datasets} for details about the datasets). As these datasets are inherently multilabel, we uniformly sample positive labels to generate training data that fits our retrieval framework. The classification performance on these datasets has been highly optimized through extensive research in the past decade. We would like to emphasize that our aim is to not obtain state-of-the-art results on these datasets but to rather verify two aspects: (i) SNM  performs better than negative sampling, and (ii) SNM is practical for large-scale deep learning. For all our experiments, we use top-$k$ variant of SNM.

\textbf{Model architecture.}
As mentioned earlier, a simple model is used in our experiments to support our theoretical result. Our model is based on a simple embedding based neural network. For each data point $(x, y)$ in the data set, $x \in \mathbb{R}^d$, which is typically sparse, is first embedded into $512$-dimensional vector space using a two layer neural network with layer sizes $512$ and $512$ i.e., the embedding is obtained by first multiplying with a $d \times 512$ weight matrix followed by ReLU activation function, and then multiplying by a $512 \times 512$ linear transformation. The embedding is finally normalized so that its $l_2$-norm is 1. This yields a $512$-dimensional embedding representation of the input. We found including the linear layer helped accelerate training when using SGD. Each class is represented as a $512$-dimensional normalized vector. The number of parameters in this setup is $512 \cdot (d + 512 + K)$. The score of a data point is obtained by computing the inner product between the feature and class embeddings. Since all the embeddings are normalized, scores lie in $[-1, 1]$ interval.

\begin{table*}[t]
\centering
\begin{tabular}{ccccc|cccccc}
\multicolumn{2}{c}{}
& \multicolumn{3}{c}{\bf Embedding-based}
& \multicolumn{4}{c}{\bf Other Methods} \\
& & Ours & SLEEC & LEML & PfastreXML & DiSMEC & PD-Sparse & PPD-Sparse\\
\hline
& P@1 & 81.58 & 90.53 & - & 91.75 & 93.4 & 90.60 & - \\
\amcat & P@3 & 71.54 & 76.33 & - & 77.97 & 79.1 & 75.14 & - \\
& P@5 & 58.79 & 61.52 & - & 63.68 & 64.1 & 60.69 & - \\
\hline
& P@1 & 60.65 & 54.83 & 19.82 & 56.05 & 64.4 & 61.26 & 64.08 \\
\lshtc & P@3 & 42.08 & 33.42 & 11.43 & 36.79 & 42.5 & 39.48 & 41.26\\
& P@5 & 31.87 & 23.85 & 8.39 & 27.09 & 31.5 & 28.79 & 30.12 \\
\hline
& P@1 & 44.68 & 35.05 & 8.13 & 39.46 & 44.7 & - & 45.32 \\
\amsmall & P@3 & 40.55 & 31.25 & 6.83 & 35.81 & 39.7 & - & 40.37\\
& P@5 & 37.40 & 28.56 & 6.03 & 33.05 & 36.1 & - & 36.92 \\
\hline
\end{tabular}
\caption{Performance comparison with other methods on \amcat, \lshtc and \amsmall. Although our goal is to optimize Recall@$k$, we report Precision@$k$ here for comparison since it has been more widely reported in related works. We note that our embedding-based model trained using SNM performs significantly better than prior embedding-based methods such as SLEEC and LEML ~\citep{sleec, Yu14} on large datasets. Furthermore, despite its simplcity, our method is competitive with other computationally expensive methods specifically developed for these datasets.}
\label{tb:pre-table}
\end{table*}

\textbf{Training setup.} Experiments are conducted under the ``BOWL'' setting with hinge loss. We observed similar behavior for the ``POWL'' setting. SGD with a large learning rate is used in optimizing the embedding layers, and SGD with momentum is used in optimizing the linear transformation. For the small \amcat, the size of the sampled size is set as $B=1,024$, and for all other datasets, we use $B = 32,768$.  These values of $B$ are selected based on computational and memory constraints. Increasing the value of $B$ in \amcat, did not result in any significant gain in performance.

We compare different settings of the top $k$ negatives in stochastic negative mining (Table~\ref{tb:topk}). Each of the dataset comes with a pre-defined train/test split and the results we report here are based on the test data. Although the goal of the paper is to optimize Recall@$k$, we also report the Precision@$k$ metric in Table~\ref{tb:topk} since it has been more widely reported in related works. Note that the values in Table~\ref{tb:topk} are normalized with the value of negative sampling to enable easy comparison. Thus, any value greater than $1$ indicates better performance in comparison to negative sampling. The results demonstrate that top-$k$ SNM with various values of $k$ substantially improves over negative sampling, and moreover, using more aggressive mining, i.e., smaller $k$, improves the results. For all the datasets, the best result is achieved with $k = 1$. Also note that SNM does not incur any additional computational cost in comparison to negative sampling approach; in fact, it is slightly more efficient due to the fact that fewer backpropagations are needed compared to negative sampling.

In Table~\ref{tb:pre-table}, we also compare our results with a few other recent works including  SLEEC ~\citep{sleec}, LEML ~\citep{Yu14},  PfastreXML ~\citep{JPV16}, DiSMEC ~\citep{BS17} and PPD-Sparse ~\citep{YHR+17} on \amsmall, \amcat, and \lshtc. As noted earlier, the goal of our experiments is not to achieve state-of-the-art results but to opt for a simple neural network model and verify that our proposed SNM method outperforms negative sampling. However, despite its simplicity, our method is better than other embedding based methods like SLEEC, LEML, and competitive with many recently published works, including large sparse linear models where no low-rank assumptions are made. We believe that the proposed technique can be combined with more sophisticated neural network models to further improve the performance.

\section{DISCUSSION}

In this paper, we considered the problem of retrieving the most relevant classes for any given input in the specific setting of large output spaces. We provided a family of loss functions that satisfy various desirable properties for this setting, and gave a scalable technique, Stochastic Negative Mining, that can optimize instances of losses in this family. We analyzed the generalization performance of models trained using the losses in this family. Our theoretical results indicate that the Top-$k$ variant of Stochastic Negative Mining should be particularly favorable to this setting, and indeed comprehensive experiments on large public datasets indicate that this form of Stochastic Negative Mining yields substantial benefits over commonly used negative sampling techniques.

The most intriguing direction for future work is combining SNM with a custom optimization method designed to exploit the specific structure of the loss function. In particular, a principled approach to change the number of sampled classes as the optimization proceeds is an important future work. Also, here we mainly focused on a particular variant of SNM called top-$k$ SNM. It is an interesting direction of future work to investigate other settings of $\vartheta$ parameters for SNM within the sampled classes. Finally, SNM approaches for coupled loss functions such as softmax cross-entropy remains open and is left as future work.

\bibliographystyle{abbrvnat}
\bibliography{bibfile}

\appendix
\onecolumn
\section{Appendix}

\subsection*{Appendix Notation}
We use $\Delta_K$ to denote the subset of probability simplex i.e.,
$$
\Delta_K \subset \left\{\alpha \in \mathbb{R}^K \Big| \sum_{i=1}^K \alpha_i = 1, \alpha_i \geq 0 \right\}.
$$
Let $\Psi_\ell(\alpha, v) = \sum_{i=1}^K \alpha_i \ell(v, i)$. For the ease of exposition, we define the following function:
$\Psi_\ell^*(\alpha) := \inf_{v \in \Omega} \Psi_\ell(\alpha, v)$. We use $\ell_b: \Delta_K \times \mathcal{S} \rightarrow \mathbb{R}^+ \cup \{0\}$ to denote the following function:
$
\ell_b(\alpha, S) = \sum_{i \in [K] \backslash S} D(Y = i \mid X).
$

\section{Proof of Theorem~\ref{thm:calibration}}
\label{sec:asymptotic-convergence-proof}

\begin{proof}
We generalize the result in \cite{Zhang04} for our proof. For the sake of clarity, we use $\alpha$ to denote the vector $[D(Y=1|X=x), \cdots, D(Y=K|X=x))]$. We first state few definitions and auxiliary results required for the proof. We define the following function:
  $$
  \Delta R _{\ell_b, \Psi_\ell}(\epsilon) = \inf \left\{\Psi_\ell(\alpha, v) - \inf_{v \in \Omega}\Psi_\ell(\alpha) \mid \ell_b(\alpha, \topk(v)) - \inf_{v \in \Omega} \ell_b(\alpha, \topk(v)) \geq \epsilon \right\} \cup \{+\infty\}.
  $$
  The main idea of the proof is to show that $\Delta R_{\ell_b, \Psi_\ell}(\epsilon) > 0$ for $\epsilon > 0$. This essentially proves that the excess risk based on surrogate loss is non-zero whenever the excess Bayes risk is non-zero, also providing a bound on excess Bayes risk based on excess surrogate risk. Corollary 26 of \cite{Zhang04}, stated below, formalizes this intuition.  

  \begin{Lem}[\cite{Zhang04}] Suppose function $\ell_b(\alpha, \topk(v))$ is bounded and $\Delta R_{\ell_b, \Psi_\ell} > 0$ for all $\epsilon > 0$, then there exists a concave function $\xi$ on the domain $[0, +\infty]$ that depends only on $\ell_b$ and $\Psi_\ell$ such that $\xi(0) = 0, \lim_{\epsilon \rightarrow 0^+} \xi(\epsilon) = 0$ and we have
    $$
    R(h) - \inf_{h' \in \mathcal{H}}R(h') \leq \xi(R_\ell(h) - \inf_{h' \in \mathcal{H}}R_{\ell}(h'))
    $$
\label{lem:thm1-inter-lem1}
\end{Lem}

In order to show $\Delta R_{\ell_b, \Psi_\ell}(\epsilon) > 0$ for all $\epsilon > 0$, we need the following result. This follows as a modification of Lemma 28 in \cite{Zhang04} and is only included here for the sake of clarity.
\begin{Lem}
$\forall \epsilon > 0$, $\exists \delta > 0$ such that $\forall \alpha \in \Delta_K$:
$$
\inf \left\{ \Psi_\ell(\alpha, v): v_i \leq v_{[k]} \leq v_j, \alpha_j \leq \alpha_{[k]} \leq \alpha_i, \alpha_j \leq \alpha_i - \epsilon \right\} \geq \Psi_\ell^*(\alpha) + \delta.
$$
\label{lem:thm1-inter-lem2}
\end{Lem}
\begin{proof}
The proof is similar to Lemma 28 of \cite{Zhang04} except for the modification that the infimum is over the set $\{v \in \Omega \mid v_i \leq v_{[k]} \leq v_j, \alpha_j \leq \alpha_{[k]} \leq \alpha_{i}, \alpha_j \leq \alpha_i - \epsilon\}$.
\end{proof}

To prove Theorem~\ref{thm:calibration}, we observe the following: Suppose $\ell_b(\alpha, \topk(v)) \geq \inf_{v \in \Omega}\ell_b(\alpha, \topk(v)) + \epsilon$ for some $v \in \Omega$ and $\alpha \in \Delta_K$, then there $\exists i$ such that $v_i \geq v_{[k]}$ and $\alpha_{i} \leq \alpha_{[k]} - \epsilon$. To show this, we observe the following:
\begin{align*}
  \ell_b(\alpha, \topk(v)) &= 1 - \sum_{j \in \topk(v)} \alpha_j \geq \inf_{v \in \Omega} \ell_b(\alpha, \topk(v)) + \epsilon \geq 1 - \sum_{j=1}^k \alpha_{[j]} + \epsilon,
\end{align*}
and therefore, $\sum_{j \in \topk(v)} \alpha_j \leq \sum_{j=1}^k \alpha_{[j]} - \epsilon$. Note that since $|\topk(v)| = k$, from the above inequality, it is clear that there exists $i \in \topk(\alpha)$, $i \notin \topk(v)$ and $j \in \topk(v)$, $j \notin \topk(\alpha)$ such that $\alpha_j \leq \alpha_{i} - \tfrac{\epsilon}{k}$. Furthermore, From Lemma~\ref{lem:thm1-inter-lem2}, we know that $\inf \{ \Psi_\ell(\alpha, v): v_i \leq v_{[k]} \leq v_j, \alpha_i \geq \alpha_{[k]} \geq \alpha_j, \alpha_j \leq \alpha_i - \tfrac{\epsilon}{k} \} \geq \Psi_\ell^*(\alpha) + \delta$. Therefore, $\Delta R_{\ell_b, \Psi_\ell}(\epsilon) > 0$. Using Lemma~\ref{lem:thm1-inter-lem1}, we get the required result.
\end{proof}

\section{Proof of Lemma~\ref{lem:prob-dominance}}
\label{sec:powl-bowl-proof}

\begin{proof}
The fact that $\ellsnm$ is a POWL or BOWL is evident from the formula for the random variable $L(v, y)$.

Let $\sigma$ be a permutation of $[K-1]$ which sorts the coordinates of $v^{-y}$ in non-increasing order, i.e. $v^{-y}_{\sigma(j)} \geq v^{-y}_{\sigma(j')}$ for $j < j'$. Then we have
Note that
\begin{align}
	\theta_j &= \E_{\mathcal{B}}\left[\sum_{i=1}^B \vartheta_i \ind(\sigma(j) \in \mathcal{B} \text{ and } v_{\sigma(j)} \text{ is the } i^\text{th} \text{ largest score in } \mathcal{B})\right] \notag \\
	&=\ \sum_{i=1}^B \vartheta_i \Pr_{\mathcal{B}}[\sigma(j) \in \mathcal{B} \text{ and } v_{\sigma(j)} \text{ is the } i^\text{th} \text{ largest score in } \mathcal{B}] \notag \\
	&=\ \sum_{i=1}^B \vartheta_i \Pr_{\mathcal{B}}[v_{\sigma(j)} \text{ is the } i^\text{th} \text{ largest score in } \mathcal{B} | \sigma(j) \in \mathcal{B}] \cdot \frac{B}{K-1}. \label{eq:theta-formula}
\end{align}
Now, if $j' > j$, then since $v_{\sigma(j')} \leq v_{\sigma(j)}$, we have
\[Pr_{\mathcal{B}}[v_{\sigma(j)} \text{ is the } i^\text{th} \text{ largest score in } \mathcal{B} | \sigma(j) \in \mathcal{B}] \geq Pr_{\mathcal{B}}[v_{\sigma(j')} \text{ is the } i^\text{th} \text{ largest score in } \mathcal{B} | \sigma(j') \in \mathcal{B}]. \]
This is easy to check by comparing the two events. Since the coordinates of $\vartheta$ are non-increasing, this implies that $\theta_j \geq \theta_{j'}$, thus establishing that the coordinates of $\theta$ are also non-increasing.

Next, suppose that $\vartheta_i = \frac{K-1}{kB}$ for $i \in [k]$. Let $j \in [k]$. Note that if $\sigma(j) \in \mathcal{B}$, then $v_{\sigma(j)}$ is among the top $k$ scores in $\mathcal{B}$. Thus by \eqref{eq:theta-formula}, we conclude that $\theta_j = \nicefrac{1}{k}$.

Finally, if $\vartheta_i > 0$ for all $i \in [B]$, then by \eqref{eq:theta-formula}, we have $\theta_j > 0$.



\end{proof}

\section{Proofs of Theorems~\ref{thm:convexity} and \ref{thm:surrogate}}
\label{sec:convexity-surrogate-proofs}

\begin{proof}[Proof of Theorem~\ref{thm:convexity}]
Consider the $(\phi, \theta)$-POWL $\ell$. Fix any class $y \in \Yc$. Since $\phi$ is a non-increasing function, we have $\phi(v_y - v^{-y}_{[j]}) \geq \phi(v_y - v^{-y}_{[j']})$ if $j < j'$. Since $\theta$ has non-increasing coordinates, by the Rearrangement Inequality, we conclude that for any permutation $\sigma$ of $[K-1]$, we have
\[ \sum_{j=1}^{K-1} \theta_{\sigma(j)}\phi(v_y - v^{-y}_{j}) \leq  \sum_{j=1}^{K-1} \theta_{j}\phi(v_y - v^{-y}_{[j]}) = \ell(v, y). \]
Since the above inequality holds for any permutation $\sigma$, we have
\[ \ell(v, y)= \max_\sigma \sum_{j=1}^{K-1} \theta_{\sigma(j)}\phi(v_y - v^{-y}_{j}).\]
Note that $v \mapsto \sum_{j=1}^{K-1} \theta_{\sigma(j)}\phi(v_y - v^{-y}_{j})$ is a convex function of $v$ since it is non-negative linear combination of convex functions of $v$. Hence $\ell(v, y)$ is a convex function of $v$ since it is the maximum of convex functions of $v$. 

The proof that the $(\phi, \theta)$-BOWL is also convex is very similar and is omitted for brevity.
\end{proof}

\begin{proof}[Proof Theorem~\ref{thm:surrogate}]
First, consider the $(\phi, \theta)$-POWL $\ell$. Suppose $y \not\in \topk(v)$. Then for any $j \in [k]$, we have $v_y \leq v^{-y}_{[j]}$, and so $\phi(v_y - v^{-y}_{[j]}) \geq \ind(v_y - v^{-y}_{[j]} \leq 0) = 1$. Since $\theta$ is a non-negative vector and $\phi$ is also non-negative, we have 
\[\ell(v, y) \geq \sum_{j=1}^k \theta_j \phi(v_y - v^{-y}_{[j]}) \geq \sum_{j=1}^k \theta_j \cdot 1 = \ind(y \not\in \topk(v)).\]
If $y \in \topk(v)$, then $\ind(v \not\in \topk(v)) = 0$, and $\ell(v, y) \geq \ind(v \not\in \topk(v))$ since $\ell(v, y)$ is always non-negative.

Now, consider the $(\phi, \theta)$-BOWL $\ell$. We have
\[\ell(v, y) = \phi(v_y) + \sum_{j=1}^{K-1} \theta_j \phi(v_y - v^{-y}_{[j]}) \geq \sum_{j=1}^{k} \theta_j(\phi(v_y) + \phi(-v^{-y}_{[j]})) \geq \sum_{j=1}^{k} 2\theta_j(\phi(\tfrac{1}{2}(v_y - v^{-y}_{[j]}))).\]
The first inequality above follows since $\theta_j = \nicefrac{1}{k}$ for $j \in [k]$ and the fact that $\phi$ is always non-negative, and the second inequality by the convexity of $\phi$. Now arguing just like in the POWL case, we have
\[\sum_{j=1}^{k} 2\theta_j(\phi(\tfrac{1}{2}(v_y - v^{-y}_{[j]}))) \geq 2\ind(v \not\in \topk(v)).\]

\end{proof}

\section{Proof of Theorem~\ref{thm:snm-calibration}}
\label{sec:calibration-proof}

\begin{proof}
We first prove the following key order-preserving property of the loss functions in Definition~\ref{def:snm-pairwise-loss-alt} and ~\ref{def:snm-binary-loss-alt} (the proof of the result is given in Lemma~\ref{lem:snm-pairwise-order} and Lemma~\ref{lem:snm-binary-order}).
\begin{Lem*}
Suppose $\phi$ satisfies the conditions in Theorem~\ref{thm:snm-calibration}. Then for any $\alpha \in \Delta_K$ that satisfies the following condition:
$$
\alpha_{[k]}> \frac{\sum_{l=k+1}^{k+q} \alpha_{[l]}}{k \sum_{j=k}^{k+q-1} \theta_{j} },
$$
for all $q \in [K-k]$ and $v \in \mathbb{R}^K$ such that $\Psi_\ell(\alpha, v) = \Psi_\ell^*(\alpha)$ for $\ell$ in Definition~\ref{def:snm-pairwise-loss-alt} and Definition~\ref{def:snm-binary-loss-alt} with appropriate conditions on $\{\theta_i\}_{i=1}^{K-1}$ (as specified in Theorem~\ref{thm:snm-calibration}), we have
\begin{enumerate}
\item $v_i \geq v_j$ when $\alpha_i > \alpha_j$ and
\item $v_{[i]} > v_{[j]}$ when $\alpha_i > \alpha_j$ and $i \in [k]$ and $j \in [K] \backslash [k]$.
\end{enumerate}
\end{Lem*}
The proof can be completed by appealing to the order preserving property of $\Psi_\ell$ in the above lemma. In particular, consider $v'$ such that $\Psi_{\ell}(\alpha, v') = \Psi^*_\ell(\alpha)$, then it is shown that $v_{[i]} > v_{[j]}$ when $\alpha_i > \alpha_j$ and $i \in [k]$ and $j \in [K] \backslash [k]$. From this result, it is easy to see that $\lim_{t \rightarrow \infty} \Psi_\ell(\alpha, v^t) = \Psi_\ell(\alpha, v) > \Psi_\ell(\alpha, v')= \inf_{v \in \Omega} \Psi_\ell(\alpha, v) = \Psi_\ell^*(\alpha)$, thus, completing the proof.
\end{proof}

\subsection{Lemmatta for Theorem~\ref{thm:snm-calibration}}

\begin{Lem}
Suppose $\phi$ satisfies the conditions in Theorem~\ref{thm:snm-calibration}. Then for any $\alpha \in \Delta_K$ that satisfies the following condition:
$$
\alpha_{[k]}> \frac{\sum_{l=k+1}^{k+q} \alpha_{[l]}}{k \sum_{j=k+1}^{k+q} \theta_{j} },
$$
for all $q \in [K-k]$ and $v \in \mathbb{R}^K$ such that $\Psi_\ell(\alpha, v) = \Psi_\ell^*(\alpha)$ for $\ell$ in Definition~\ref{def:snm-pairwise-loss-alt} with $\theta_j = \frac{1}{k}$ for all $j \in [k]$ and $\theta_j \leq \tfrac{1}{k}$ for $j > k$, we have
\begin{enumerate}
\item $v_i \geq v_j$ when $\alpha_i > \alpha_j$ and
\item $v_{[i]} > v_{[j]}$ when $\alpha_i > \alpha_j$ and $i \in [k]$ and $j \in [K] \backslash [k]$.
\end{enumerate}
\label{lem:snm-pairwise-order}
\end{Lem}
\begin{proof}
We prove the first part by contradiction. Assume $\exists j_1, j_2$ such that $\alpha_{j_1} > \alpha_{j_2}$ but $v_{j_1} < v_{j_2}$. Consider $\bar{v}$ such that $\bar{v}_i = v_i$ for all $i \neq j_1, j_2$, $\bar{v}_{j_1} = v_{j_2}$ and $\bar{v}_{j_2} = v_{j_1}$. Then we have
\begin{align*}
&\Psi_\ell(\alpha, \bar{v}) - \Psi_\ell(\alpha, v) \nonumber \\
&= \alpha_{j_1}\left(\sum_{j=1}^{K-1} \theta_j \phi(\bar{v}_{j_1} - \bar{v}^{-j_1}_{[j]}) - \sum_{j=1}^{K-1} \theta_j \phi(v_{j_1} - v^{-j_1}_{[j]}) \right) + \alpha_{j_2} \left(\sum_{j=1}^{K-1} \theta_j \phi(\bar{v}_{j_2} - \bar{v}^{-j_2}_{[j]}) - \sum_{j=1}^{K-1} \theta_j \phi(v_{j_2} - v^{-j_2}_{[j]}) \right) \nonumber \\
&= (\alpha_{j_1} - \alpha_{j_2})\left(\sum_{j=1}^{K-1} \theta_j \phi(v_{j_2} - v^{-j_2}_{[j]}) - \sum_{j=1}^{K-1} \theta_j \phi(v_{j_1} - v^{-j_1}_{[j]})\right)
\end{align*}
The above equality is due to the definition of $\bar{v}$. Furthermore, we observe the following: $v_{j_2} > v_{j_1}$ and $v^{-j_1}_{[j]} \geq v^{-j_2}_{[j]}$ for all $j \in [K-1]$. This is due to the fact that removal of $v_{j_2}$ rather than $v_{j_1}$ from $v$ can only decrease the order statistic.. Therefore, we have
$$
v_{j_2} - v^{-j_2}_{[j]} > v_{j_1} - v^{-j_1}_{[j]},
$$
for all $j \in [K-1]$. Since $\phi$ is non-increasing, it is clear that $\Psi_\ell(\alpha, \bar{v}) - \Psi_\ell(\alpha, v) \leq 0$ . Also, note that at least one $v_{j_1} - v^{-j_1}_{[j]} < 0$ since $v_{j_2} > v_{j_1}$ for $j \in [k]$. Since $\phi$ is strictly decreasing on $(-\infty, 0]$, we can, in fact, obtain $\Psi_\ell(\alpha, \bar{v}) - \Psi_\ell(\alpha, v) < 0$, which is a contradiction to the optimality of $v$.

We now focus on the second part of the proof. Without loss of generality, suppose $\alpha_1 \geq \cdots \geq \alpha_k > \alpha_{k+1} \geq \cdots \alpha_K$.  Suppose $v_{k} > v_{k+1}$, then the second part follows immediately. Now, consider the scenario:
$$
v_1 \geq v_2 \geq \cdots \geq v_{k} = v_{k+1} = \cdots = v_{k+q} > v_{k+q+1} \geq \cdots \geq v_K.
$$
We will prove that such a scenario is not possible. We prove this by contradiction. Consider the vector $v'$ defined as follows:
\[
v'_i= 
\begin{cases}
v_i + \delta, & \text{for } i = k \\
v_i - \beta \delta, & \text{for } k+1 \leq i \leq k+q \\
v_i, & \text{otherwise }.
\end{cases}
\]
Here $\delta$ is chosen sufficiently small such that $v'_{k+q} > v'_{k+q+1}$ with $\beta = \tfrac{1}{k\sum_{j=k+1}^{k+q}\theta_j}$. When $\alpha, v$ are held fixed, with slight abuse of notation, we use $\Psi_\ell(\delta)$ to denote part of the function $\Psi_\ell(\alpha, v')$ that only depends on $\delta$. Let us denote the remaning part by $C_{\alpha, v}$ such that $\Psi_\ell(\alpha, v) = \Psi_\ell(0) + C_{\alpha, v}$. More specifically, we have the following:
\begin{align*}
\Psi_\ell(\delta) &= \underbrace{\alpha_k \left[\sum_{j=1}^{k-1} \frac{1}{k} \phi(v_{k} - v_j + \delta) + \sum_{j=k+q+1}^K \theta_{j-1}\phi(v_k - v_j + \delta) + \sum_{j=k+1}^{k+q} \theta_{j-1}\phi(v_{k} - v_{j} + (1+\beta) \delta)\right]}_{T_1(\delta)} \\
& \quad + \underbrace{\sum_{l=k+1}^{k+q} \alpha_l \left[\sum_{j=1}^{k-1} \frac{1}{k}\phi(v_{l} - v_j -\beta \delta) + \sum_{j=k+q+1}^K \theta_{j-1}\phi(v_l - v _j - \beta\delta) + \frac{1}{k}\phi(v_{l} - v_k - (1+\beta) \delta) \right]}_{T_2(\delta)} \\
& \quad + \underbrace{\sum_{l=1}^{k-1} \alpha_l \left[\frac{1}{k} \phi(v_{l} - v_k -\delta) + \sum_{j=k+1}^{k+q} \theta_{j-1}\phi(v_l - v _j + \beta\delta)\right]}_{T_3(\delta)} \\
& \quad + \underbrace{\sum_{l=k+q+1}^{K} \alpha_l \left[\frac{1}{k} \phi(v_{l} - v_k -\delta) + \sum_{j=k+1}^{k+q} \theta_{j}\phi(v_l - v _j + \beta\delta)\right]}_{T_4(\delta)}
\end{align*}
Since, $\theta_{i} = 1/k$ for all  $i \leq k$, $\Psi_\ell(\alpha, v') = \Psi_\ell(\delta) + C_{\alpha, v}$ for $0 \leq \beta\delta \leq v'_{k+q} - v'_{k+q+1}$. This follows the fact the the rank (position when sorted) of of $v'_i$  amongst elements in $v'$ is same as that of $v_i$ amongst elements in $v$ for $i > k$ for sufficiently small chosen $\delta$ since  the rank of $v'_k$ in $v'$ can only decrease in comparison to rank $v_k$ in $v$ and the rank remains same for all $i > k$. Also, note that $\Psi_\ell$ is differentiable. Our aim is to show that $\Psi'_\ell(0) < 0$, which implies a contradiction to the optimality of $v$. To this end, we analyze the differential of aforementioned terms separately as follows:
\begin{align*}
T'_3(\delta) = \sum_{l=1}^{k-1} \alpha_l \left[-\frac{1}{k}\phi'(v_{l} - v_k -\delta) + \beta \sum_{j=k+1}^{k+q} \theta_{j-1}\phi'(v_l - v _j + \beta\delta)\right] \\
= \sum_{l=1}^{k-1} \alpha_l \left[-\frac{1}{k}\phi'(v_{l} - v_k -\delta) + \phi'(v_l - v _k + \beta\delta) \beta \sum_{j=k+1}^{k+q} \theta_{j-1}\right].
\end{align*}
The above equality holds because $v_k = v_{i}$ for all $i \in [k+1, k+q]$. From the above equality, we have:
\begin{align*}
T'_3(0) = \sum_{l=1}^{k-1} \alpha_l \left[\left(\beta \sum_{j=k+1}^{k+q} \theta_{j-1}-\frac{1}{k}\right)\phi'(v_{l} - v_k)\right] \leq 0.
\end{align*}
This is due to the fact that $\phi$ is non-increasing and following inequality
$$
\beta \sum_{j=k+1}^{k+q} \theta_{j-1} \geq \frac{1}{k}.
$$
In a similar manner, it can also be shown that $T'_4(0) = 0$. To complete the proof, we need to show that $T'_1(0) + T'_2(0) < 0$. We observe the following:
\begin{align*}
T'_1(\delta) + T'_2(\delta) &= \frac{1}{k} \sum_{j=1}^{k-1} \left(\alpha_k\phi'(v_{k} - v_j + \delta) - \phi'(v_{k} - v_j -\beta \delta) \beta \sum_{l=k+1}^{k+q} \alpha_l \right) \\
 &\quad + \sum_{j=k+q+1}^K \theta_{j-1}\left(\alpha_k \phi'(v_k - v_j + \delta) - \phi'(v_{k} - v_j -\beta \delta) \beta \sum_{l=k+1}^{k+q} \alpha_l \right) \\
& \quad + \phi'((1+\beta) \delta) (1+\beta)\alpha_k\sum_{j=k+1}^{k+q} \theta_{j-1} - \frac{1 + \beta}{k} \phi'(-(1+\beta) \delta) \sum_{l=k+1}^{k+q} \alpha_l 
\end{align*}
The above equality is due to the fact that $v_k = v_{i}$ for all $i \in [k+1, k+q]$. From the above equality we have,
\begin{align*}
T'_1(0) + T'_2(0) &= \frac{1}{k} \sum_{j=1}^{k-1}\phi'(v_{k} - v_j) \left(\alpha_k - \beta \sum_{l=k+1}^{k+q} \alpha_l \right) \\
 &\quad + \sum_{j=k+q+1}^K \phi'(v_k - v_j) \theta_{j-1}\left(\alpha_k  - \beta \sum_{l=k+1}^{k+q} \alpha_l \right) \\
& \quad + (1+\beta)\phi'(0)\left[\alpha_k\sum_{j=k+1}^{k+q} \theta_{j-1} - \frac{1}{k}\sum_{l=k+1}^{k+q} \alpha_l\right]
\end{align*}
From the above equality, we can see that $T'_1(0) + T'_2(0) < 0$. This is due to the fact that $\phi$ is non-increasing with $\phi'(0) < 0$ and the following inequalities:
\begin{align*}
&\alpha_k > \beta \sum_{l=k+1}^{k+q} \alpha_l = \frac{\sum_{l=k+1}^{k+q} \alpha_l}{k \sum_{l=k+1}^{k+q}\theta_j}  \\
&\alpha_k\sum_{j=k}^{k+q-1} \theta_{j} > \frac{1}{k}\sum_{l=k+1}^{k+q} \alpha_l.
\end{align*}
Therefore, we have $\Psi'_\ell(0) = T'_1(0) + T'_2(0) + T'_3(0) + T'_4(0) < 0$. This is a contradiction to the optimality of $v$. Hence, the scenario 
$$
v_1 \geq v_2 \geq \cdots \geq v_{k} = v_{k+1} = \cdots = v_{k+q} > v_{k+q+1} \geq \cdots \geq v_K,
$$
is not possible. This completes the proof of second part of the lemma.
\end{proof}

\begin{Lem}
Suppose $\phi$ satisfies the conditions in Theorem~\ref{thm:snm-calibration}. Then for any $\alpha \in \Delta_K$ that satisfies the following condition:
$$
\alpha_{[k]}> \frac{\sum_{l=k+1}^{k+q} \alpha_{[l]}}{k \sum_{j=k+1}^{k+q} \theta_{j} },
$$
for all $q \in [K-k]$ and $v \in \mathbb{R}^K$ such that $\Psi_\ell(\alpha, v) = \Psi_\ell^*(\alpha)$ for $\ell$ in Definition~\ref{def:snm-binary-loss-alt} with $\theta_j = \tfrac{1}{k}$ for all $j \in [k]$ and $\theta_{j} \leq \tfrac{1}{k}$ for $j > k$, we have
\begin{enumerate}
\item $v_i \geq v_j$ when $\alpha_i > \alpha_j$ and
\item $v_{[i]} > v_{[j]}$ when $\alpha_i > \alpha_j$ and $i \in [k]$ and $j \in [K] \backslash [k]$.
\end{enumerate}
\label{lem:snm-binary-order}
\end{Lem}
\begin{proof}
We prove the first part by contradiction. Assume $\exists j_1, j_2$ such that $\alpha_{j_1} > \alpha_{j_2}$ but $v_{j_1} < v_{j_2}$. Consider $\bar{v}$ such that $\bar{v}_i = v_i$ for all $i \neq j_1, j_2$, $\bar{v}_{j_1} = v_{j_2}$ and $\bar{v}_{j_2} = v_{j_1}$. Then we have
\begin{align*}
&\Psi_\ell(\alpha, \bar{v}) - \Psi_\ell(\alpha, v) = (\alpha_{j_1} - \alpha_{j_2})\left(\phi(v_{j_2}) + \sum_{j=1}^{K-1} \theta_j \phi(- v^{-j_2}_{[j]}) - \phi(v_{j_1}) - \sum_{j=1}^{K-1} \theta_j \phi(- v^{-j_1}_{[j]})\right) 
\end{align*}
The above equality is due to the definition of $\bar{v}$. Furthermore, we observe the following: $v_{j_2} > v_{j_1}$ and $v^{-j_1}_{[j]} \geq v^{-j_2}_{[j]}$ for all $j \in [K-1]$. This is due to the fact that removal of $v_{j_2}$ rather than $v_{j_1}$ from $v$ can only decrease the order statistic. If $v_{j_1}$ is non-positive, then $\phi(v_{j_2}) < \phi(v_{j_1})$ and $\phi(-v^{-j_2}_{[j]}) \leq \phi(-v^{-j_1}_{[j]})$ as $\phi'(\epsilon) < 0$ for all $\epsilon \leq 0$ and $\phi$ is non-increasing,  which is a contradiction to the optimality of $v$. 

We now consider the case where $v_{j_2} > v_{j_1} > 0$. It is not hard to see that $v^{-j_1}_{[j]} = v^{-j_2}_{[j]}$ whenever $v^{-j_1}_{[j]} < v_{j_1}$. Furthermore, $\sum_{i=1}^{K-1} v^{-j_1}_{[j]} > \sum_{i=1}^{K-1} v^{-j_2}_{[j]}$. From the above two facts, we get $v^{-j_1}_{[j]} > v^{-j_2}_{[j]}$ for some $j$ such that $v^{-j_2}_{[j]} > 0$. For this $j$, $\phi(-v^{-j_2}_{[j]}) \leq \phi(-v^{-j_1}_{[j]})$ as $\phi'(\epsilon) < 0$ for all $\epsilon \leq 0$. Since $\phi$ is strictly decreasing on $(-\infty, 0]$, we can, in fact, obtain $\Psi_\ell(\alpha, \bar{v}) - \Psi_\ell(\alpha, v) < 0$, which is again a contradiction to the optimality of $v$. This completes the first part of the proof.

We now turn our attention to the second part.  For the ease of exposition, suppose $\alpha_1 \geq \cdots \geq \alpha_k > \alpha_{k+1} \geq \cdots \alpha_K$.  The proof is along similar lines as that of pairwise comparison method. Suppose $v_{k} > v_{k+1}$, then the second part follows immediately. Now, consider the scenario:
$$
v_1 \geq v_2 \geq \cdots \geq v_{k} = v_{k+1} = \cdots = v_{k+q} > v_{k+q+1} \geq \cdots \geq v_K.
$$
We will prove that is not possible through proof by contradiction. Consider the vector $v'$ defined as follows:
\[
v'_i= 
\begin{cases}
v_i + \delta, & \text{for } i = k \\
v_i - \beta \delta, & \text{for } k+1 \leq i \leq k+q \\
v_i, & \text{otherwise },
\end{cases}
\]
where $\delta$ is chosen sufficiently small such that $v'_{k+q} > v'_{k+q+1}$ with $\beta = \tfrac{1}{k\sum_{j=k+1}^{k+q}\theta_j}$. When $\alpha, v$ are held fixed, with slight abuse of notation, we use $\Psi_\ell(\delta)$ to denote part of the function $\Psi_\ell(\alpha, v')$ that only depends on $\delta$. Let us denote the remaning part by $C_{\alpha, v}$ such that $\Psi_\ell(\alpha, v) = \Psi_\ell(0) + C_{\alpha, v}$. More specifically, we have the following:
\begin{align*}
\Psi_\ell(\delta) &= \underbrace{\alpha_k \left[\phi(v_{k} + \delta) + \sum_{j=k+1}^{k+q} \theta_{j-1}\phi(- v_j + \beta \delta)\right]}_{T_1(\delta)} \\
& + \underbrace{\sum_{l=k+1}^{k+q} \alpha_l \left[\phi(v_{l}-\beta \delta) + \frac{1}{k} \phi(-v_k - \delta) + \sum_{j=k+1}^{l-1} \theta_{j}\phi(- v _j + \beta\delta) + \sum_{j=l+1}^{k+q}\theta_{j-1}\phi(- v_j +\beta\delta) \right]}_{T_2(\delta)} \\
& + \underbrace{\sum_{l=1}^{k-1} \alpha_l \left[\frac{1}{k}\phi(- v_k -\delta) + \sum_{j=k+1}^{k+q} \theta_{j-1}\phi(- v _j + \beta\delta)\right]}_{T_3(\delta)} + \underbrace{\sum_{l=k+q+1}^{K} \alpha_l \left[\frac{1}{k} \phi(- v_k -\delta) + \sum_{j=k+1}^{k+q} \theta_{j}\phi(- v _j + \beta\delta)\right]}_{T_4(\delta)}
\end{align*}
Since, $\theta_{i} = \frac{1}{k}$ for all  $i \leq k$, $\Psi_\ell(\alpha, v') = \Psi_\ell(\delta) + C_{\alpha, v}$ for $0 \leq \beta\delta \leq v'_{k+q} - v'_{k+q+1}$ and $\Psi_\ell(\delta)$ is differentiable as argued for POWL. Our goal is to show that $\Psi'_\ell(0) < 0$, which implies $\Psi_\ell(\alpha, v') < \Psi_\ell(\alpha, v)$, thereby contradicting the optimality of $v$. With our choice of $\beta$, it can be shown that $T'_3(0) \leq 0$ and $T'_4(0) = 0$ using the same argument for corresponding terms for POWL. To complete the proof, we need to show that $T'_1(0) + T'_2(0) < 0$. We observe the following:
\begin{align*}
T'_1(\delta) + T'_2(\delta) &= \alpha_k\phi'(v_k + \delta) + \beta\alpha_k \sum_{j=k+1}^{k+q} \theta_{j-1}\phi'(-v_k + \beta\delta) \\
& \quad + \sum_{l=k+1}^{k+q} \alpha_l \left[-\beta\phi'(v_{k}-\beta \delta) - \frac{1}{k} \phi'(-v_k - \delta) + \beta \sum_{j=k+1}^{k+q-1} \theta_{j}\phi'(- v _k + \beta\delta) \right]
\end{align*}
The above equality is due to the fact that $v_k = v_{i}$ for all $i \in [k+1, k+q]$. From the above equality we have,
\begin{align*}
T'_1(0) + T'_2(0) &= \alpha_k\phi'(v_k) + \beta\alpha_k \sum_{j=k+1}^{k+q}\theta_{j-1}\phi'(-v_k) + \sum_{l=k+1}^{k+q} \alpha_l \left[-\beta\phi'(v_{k}) - \frac{1}{k}\phi'(-v_k) + \beta \sum_{j=k+1}^{k+q-1} \theta_{j}\phi'(- v _k)  \right] \\
&= \left(\alpha_k - \beta\sum_{l=k+1}^{k+q} \alpha_l\right) \phi'(v_k) + \sum_{l=k+1}^{k+q} \left(\alpha_k\sum_{l=k+1}^{k+q} \alpha_l - \theta_{k+q} \sum_{l=k+1}^{k+q} \alpha_\ell \right) \beta \phi'(-v_k) < 0.
\end{align*}
The last inequality is due to the following:
\begin{align*}
&\alpha_k > \beta \sum_{l=k+1}^{k+q} \alpha_l = \frac{\sum_{l=k+1}^{k+q} \alpha_l}{k \sum_{l=k+1}^{k+q}\theta_j}.
\end{align*}
$\theta_{k+q} \leq \tfrac{1}{k}$ and the fact that at least one of $\phi'(-v_k)$ and $\phi'(v_k)$ is strictly negative as $\phi'(\epsilon) < 0$ for $\epsilon \leq 0$. Therefore, we have $\Psi'_\ell(0) = T'_1(0) + T'_2(0) + T'_3(0) + T'_4(0) < 0$. This is a contradiction to the optimality of $v$. Hence, the scenario 
$$
v_1 \geq v_2 \geq \cdots \geq v_{k} = v_{k+1} = \cdots = v_{k+q} > v_{k+q+1} \geq \cdots \geq v_K,
$$
is not possible. This completes the proof of second part of the lemma.
\end{proof}

\section{Proofs of Theorems~\ref{thm:owl-generalization-bound} and \ref{thm:margin-bound}}
\label{sec:gen-proofs}
\begin{proof}[Proof of Theorem~\ref{thm:owl-generalization-bound}]
Our generalization bounds are based on the work of \citet{kloft}, who give general purpose bounds in terms of Lipschitz constants and range of the loss. In particular, suppose that $|\ell(v, y)| \leq \Phi'$. Further, suppose that for any $y \in \Yc$, $\ell$ satisfies an $L_2$-Lipschitzness condition of the form:
\[|\ell(v, y) - \ell(u, y)| \leq L_1\|v - u\|_2 + L_2|v_y - u_y|,\]
and an $L_\infty$-Lipschitzness condition of the form:
\[|\ell(v, y) - \ell(u, y)| \leq L_3\|v - u\|_\infty.\]
Then \cite{kloft} prove (see Theorems 2 and 6 in their paper\footnote{While these results assume a specific linear structure of the hypothesis class, it is easy to verify that the results hold in the more general setting described here.}) that the generalization error is bounded with probability at least $1-\delta$ by
\[ \tilde{O}\left(\min\left\{L_1K\Gauss_{\bar{S}}(\mH) + L_2\Gauss_S(\mH), L_3\sqrt{K}\overline{\Rad}_{\bar{S}}(\mH)\right\}\right) + 3\Phi' \sqrt{\frac{\log(2/\delta)}{2n}}.\]
For OWLs, Lemma~\ref{lem:lipschitz} provides the required Lipschitz constants. Next, it is easy to check that the setting $\Phi' = \|\theta\|_1\Phi$ is a valid bound on the range of the losses. The claimed generalization bound follows by plugging in the values of the Lipschitz constants and $\Phi'$.
\end{proof}

\begin{Lem} \label{lem:lipschitz}
Let $\phi(\cdot)$ be $L$-Lipschitz. Let $u, v \in \R^K$ be two score vectors. Then the $(\phi, \theta)$-POWL $\ell$ satisfies the following Lipschitzness conditions, for any $y \in \Yc$:
\[|\ell(v, y) - \ell(u, y)| \leq 
\begin{cases}
L\|\theta\|_2\|v - u\|_2 + L\|\theta\|_1|v_y - u_y| & (L_2\text{-Lipschitzness}) \\
2L\|\theta\|_1\|v - u\|_\infty & (L_\infty\text{-Lipschitzness})
\end{cases}
\]
Furthermore, the $(\phi, \theta)$-BOWL $\ell$ satisfies the following Lipschitzness conditions, for any $y \in \Yc$:
\[|\ell(v, y) - \ell(u, y)| \leq 
\begin{cases}
L\|\theta\|_2\|v - u\|_2 + L|v_y - u_y| & (L_2\text{-Lipschitzness}) \\
L(\|\theta\|_1+1)\|v - u\|_\infty & (L_\infty\text{-Lipschitzness})
\end{cases}
\]
\end{Lem}
\begin{proof}
We first consider the $(\phi, \theta)$-POWL $\ell$. Let $p \in \{2, \infty\}$. Then we have
\begin{align*}
	|\ell(v, y) - \ell(u, y)| &= \left|\sum_{j=1}^{K-1} \theta_j (\phi(v_y - v_{[j]}^{-y}) - \phi(u_y - u_{[j]}^{-y}))\right| \\
	&\leq \sum_{j=1}^{K-1} \theta_j \cdot L(|v_y - u_y| + |v_{[j]}^{-y} - u_{[j]}^{-y}|)\\
	&\leq L\|\theta\|_1 |v_y - u_y| + L\|\theta\|_p\|\tilde{v}^{-y} - \tilde{u}^{-y}\|_{p/(p-1)}.
\end{align*}
The first inequality above follows from the $L$-Lipschitzness of $\phi$ and the triangle inequality, and the second by H\"{o}lder's inequality. Then applying the bounds from Lemma~\ref{lem:sorted-norm-bound}, we get the claimed bounds.

The claimed bounds for the $(\phi, \theta)$-BOWL are obtained using an almost identical analysis and is omitted for brevity.
\end{proof}

\begin{Lem} \label{lem:sorted-norm-bound}
Let $u, v \in \R^K$ be two score vectors, and let $\tilde{u}, \tilde{v} \in \R^K$ be sorted versions of $u, v$ respectively with coordinates in non-increasing order. Then we have 
\[ \|\tilde{v} - \tilde{u}\|_2 \leq \|v - u\|_2 \quad \text{ and } \quad \|\tilde{v} - \tilde{u}\|_\infty \leq \|v - u\|_\infty.\]
\end{Lem}
\begin{proof}
The first inequality is an easy consequence of the Rearrangement Inequality after squaring both sides. As for the second inequality, let $\epsilon := \|v - u\|_\infty$, and let $k \in \Yc$ be any index. Then note that for any $j \in \topk(u)$, we have $v_j \geq u_j - \epsilon$, and hence $\tilde{v}_k \geq \tilde{u}_k - \epsilon$. Similarly, $\tilde{u}_k \geq \tilde{v}_k - \epsilon$. These two inequalities imply that $|\tilde{v}_k -\tilde{u}_k| \leq \epsilon$, and thus the claimed bound follows.
\end{proof}

\begin{proof}[Proof of Theorem~\ref{thm:margin-bound}]
Consider the $(\phi_{\text{ramp}, \rho}, \theta)$-POWL $\ell$ where $\theta_k = 1$ and $\theta_j = 0$ for all $j \neq k$. Thus, this loss can be rewritten as $\ell(v, y) = \phi_{\text{ramp}, \rho}(v_y - v_{[k]}^{-y})$, and hence for a given hypothesis $h$ and an example $(x, y)$, we have $\ell(h(x), y) = \phi_{\text{ramp}, \rho}(\rho_h(x, y))$. The claimed margin bound then follows by applying the bound from Theorem~\ref{thm:owl-generalization-bound} using the facts that $\|\theta\|_1 = \|\theta\|_2 = 1$, $L = \frac{1}{\rho}$, $\Phi = 1$, and $\ind[u \leq 0] \leq \phi_{\text{ramp}, \rho}(u) \leq \ind[u \leq \rho]$ for any $u \in \R$ (and in particular, for $u = \rho_h(x, y)$).
\end{proof}

\end{document}